\documentclass[letterpaper, 10 pt, conference]{ieeeconf}  %

\IEEEoverridecommandlockouts                              %

\usepackage{graphicx}
\usepackage[utf8]{inputenc}
\usepackage{amsmath}
\usepackage{hyperref}
\usepackage{bm}
\usepackage[percent]{overpic}
\usepackage{color}
\usepackage{xcolor,colortbl}
\usepackage{tabularx}
\usepackage{adjustbox}
\usepackage{array}
\usepackage{flushend} 
\newcolumntype{R}[2]{%
	>{\adjustbox{angle=#1,lap=\width-(#2)}\bgroup}%
	l%
	<{\egroup}%
}
\usepackage{cite}

\graphicspath{{./figs/}}

\definecolor{LightGreen}{rgb}{0.898039, 1, 0.8}

\begin{document}

\title{\LARGE \bf A gripper for flap separation and opening\\of sealed bags}
\author{ 
Sergi Foix,
Jaume Oriol, 
Carme Torras, and 
J\'ulia Borr\`as. 
	\thanks{The research leading to these results receives funding from European Union's Horizon Europe programme project SoftEnable (grant agreement No. 101070600) and
    project PID2023-152259OB-I00 (CHLOE-MAP)funded by MCIU/ AEI /10.13039/501100011033 and by ERDF, UE. }
	\thanks{The authors are  with Institut de Robòtica i Informàtica Industrial, CSIC-UPC,
		Llorens i Artigas 4-6, 08028 Barcelona, Spain. {\tt \{jborras, galenya, torras\}@iri.upc.edu}}%
}

\maketitle

\begin{abstract}
Separating thin, flexible layers that must be individually grasped is a common but challenging manipulation primitive for most off-the-shelf grippers. A prominent example arises in clinical settings: the opening of sterile flat pouches for the preparation of the operating room, where the first step is to separate and grasp the flaps. We present a novel gripper design and opening strategy that enables reliable flap separation and robust seal opening. This capability addresses a high-volume repetitive hospital procedure in which nurses manually open up to 240 bags per shift, a physically demanding task linked to musculoskeletal injuries. Our design combines an active dented-roller fingertip with compliant fingers that exploit environmental constraints to robustly grasp thin flexible flaps\footnote{Gripper design in the patent process under the European Union patent~\cite{foix2024Gripper}.}. Experiments demonstrate that the proposed gripper reliably grasps and separates sealed bag flaps and other thin-layered materials from the hospital, the most sensitive variable affecting performance being the normal force applied. 
When two copies of the gripper grasp both flaps, the system withstands the forces needed to open the seals robustly. To our knowledge, this is one of the first demonstrations of robotic assistance to automate this repetitive, low-value, but critical hospital task.
\end{abstract}

\IEEEpeerreviewmaketitle

\section{Introduction}
\label{sec:introduction}
Robotic manipulation is challenging because it combines high-level understanding with complex interactions with objects and environment. Besides the classic grasping where the contact occurs only between hand and object, many new approaches have explored other non-prehensile strategies such as environmental constraints~\cite{eppner2015Exploitation}, manipulation with pushing~\cite{chavan2018stable} or caging with energies~\cite{mahler2016energy}, to mention a few.

Manipulation of deformable objects goes a step further in the complexities of manipulation at all levels, from the state semantic understanding to their perception and manipulation. One of the reasons of this complexity is the necessity of fully utilizing prehensile and non-prehensile strategies to achieve successful manipulation, applying grasping and manipulation primitives way beyond the use of only the robot grippers, but fully exploiting friction, dynamics, tactile senses and the environment~\cite{johannsmeier2025process, blanco2025TDOM}.

Once identified the required manipulation primitives, smart hand designs can greatly simplify their execution by integrating passive and active mechanisms to facilitate them. In this context, rapid prototyping allows fast testing with real interactions, eliminating reliance on simulations that often differ significantly from reality, specially for such complex grasping primitives. This is specially relevant for 2D deformable objects like plastic films or clothes. Such objects have numerous unique characteristics that present considerable challenges for robotic manipulation. Their complexity, including intricate shape deformations, make simulations often fail to bridge the gap between virtual and real-world scenarios. Therefore, the ability to rapidly test manipulation primitives on actual objects becomes not only appealing but essential to exploit friction, thin layer separation, thicker edges, seams, or the use of the environment as a tool. Exploring mechanical prototypes that facilitate the exploitation of such elements allows to reach where current computer vision or learning methods fall short.

\begin{figure}[tb]
    \centering
    \includegraphics[width=\linewidth]{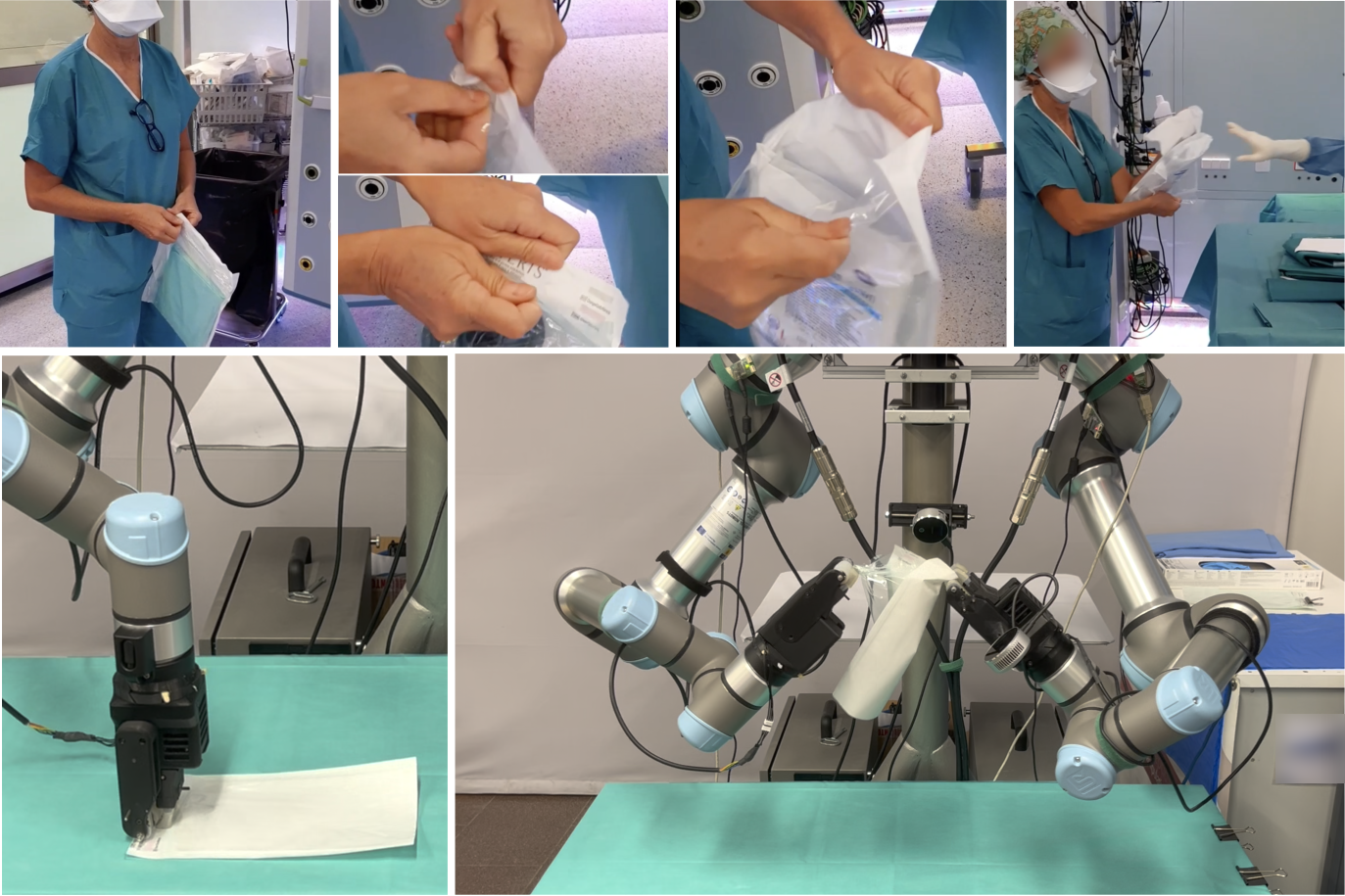}
    \caption{(Top) A nurse opening a sealed bag for the operating room preparation. (Bottom) A bi-manual robotic system, based on two UR5E manipulators, performing the same task. First grasping from the table (left) and then opening (right).}
    \label{fig:openingBags}
\end{figure}

One of these challenging tasks is that of thin layer separation. Such a problem has many industrial solutions for piles of cut cloth pieces \cite{yamazaki2021versatile, zhu2024end, unde2024design}, but solutions beyond the industrial ones are also required because this skill is needed for many daily tasks such as unfolding, opening bags or turning pages of a book, just to mention a few. In this work we present a mechanical gripper design that solves the task of layer separation, and in particular it was designed to be part of a process of opening sealed bags during the preparation of operating rooms in hospitals (Fig. \ref{fig:openingBags}-top), in the context of the European project SoftEnable. 
This task is repetitive and tiring. Each surgery requires many bag openings, performed manually by the circulating nurse, who can do between 100 and 240 openings during each shift. It is also physically demanding, as the bag contents can weigh 20g to almost 1 kg and in the long run cause musculoskeletal injuries and pain~\cite{asghari2019musculoskeletal}. Hospital protocols require that each bag is opened without touching the interior and then offered to the instrument nurse while ensuring that the sterility of the material is maintained. This is a very challenging task for robots, with high variability due to not only different weights but also bag sizes, shapes, materials, and type of seal. 
This task 
has gained attention recently in industry, with companies such as~\cite{siemensAURORA} developing a robot to do this task, among other hospital assistance. However, in their approach they assume the flaps of the bag are pre-separated and ready to be grasped by their robot, and our paper solves precisely this step.

To the authors' knowledge, this work presents the first solution to solve the separation of layers and grasping of the flaps to open bags (second and third image in the top row images in Fig. \ref{fig:openingBags}), and shows preliminary results for the whole task for the type of bags our hospital collaborators shared with us. The separation of layers required to separate and grasp the flaps entails studying the interactions, friction requirements, and exploiting other material characteristics that may occur during the grasping phases. 
The design requirements that arise for opening sealed bags are:
 \begin{itemize}
     \item The need to separate and grasp one of the flaps, and offer the second one to another gripper.
     \item Strong grip of the flap, so that it can be pulled very strongly to open the seal.
     \item Ability to work on the edge of the material, as often the flaps are short.
 \end{itemize}
These requirements can be used as inspiration for the design of different parts of the gripper, which have been thought of as finger designs that can be mounted on existing parallel grippers already integrated with a robotic arm, to ease the implementation of different grasping primitives. 

\section{Related work} \label{sec:rel_works}

In the literature, there are designs to perform the task of grabbing thin layers or layer separation, especially for handling clothing. In this section, we will revise several designs, but, since they are not designed for the task of opening seal bags, none of the existing designs would work for our task, up to the authors' knowledge.

For example,~\cite{yamazaki2021versatile} proposes a clamp and \cite{abe2020robotic,zhu2024end} variations with a roller covered in something similar to Velcro, which they use to separate the top layer in a stack of cut pieces of clothing. The principle is similar, but the proposed mechanism only works for a certain type of fabric, very thin and very porous, so that it attaches to the roller. Moreover, the grasp that occurs after separating the layers is between the rollers, which provides a weaker grip compared to our design, and similarly in~\cite{manabe2021single}. The task of separating the top layer of a stack of cut clothing has also been solved with other designs, such as \cite{digumarti2021dexterous}, which uses a finger with an electro-adhesive mechanism, and~\cite{ku2020delicate}, with micro-needles that allow the fabric to be grabbed with such precision that only the top layer is taken. In both cases, their use is demonstrated only on certain types of fabric, but it would be difficult to work on plastic. More recently,~\cite{unde2024design} solved the same task by also using a gripper with rollers, and a more extensive test with different roller surfaces, materials, or holding force. However, the design is optimized for top grasps of fabric not convenient for flap grasping, and its shape would prevent the offering of the second flap to be grasped. The design in~\cite{zhang2023modular} combines a more conventional gripper with a roller mounted on one of the fingers, which can grab clothing or plastic materials that are flat on top of a table. The roller is textured similar to ours; however, the ability to separate layers is not demonstrated, but only that of lifting the clothing to be able to grab it from the table.

Other works solve similar tasks in other contexts. For instance,~\cite{zhao2023flipbot} can turn the pages of a book or magazine with a simple mechanism based on friction and adaptation to contact. This mechanism is very successful in turning pages, but grabs the lifted layer by slightly bending the material, enough to lift a piece of paper but not to hold the layer. Due to the different stiffness of the materials, it would not work for textiles or soft plastics.

The operation of opening bags, as such, is starting to be considered in the robotics literature, but has not been widely studied yet. In~\cite{chen2023bagging}, a sensorized gripper is shown to be capable to detect whether it has grabbed 0, 1, or 2 layers. The gripping of the layers is done in the air, assuming that the bag is in a crumpled form (not flat) on a table and has already been opened, at least partially.

\section{Problem statement}
We define the problem of layer separation as the one of grasping a thin layer of material that is laying on top of one or many other layers of thin material, not necessarily the same. In addition, we also assume that the pile of layers is on a table surface.

\begin{figure}[t]
    \centering
    \includegraphics[width=0.9\linewidth]{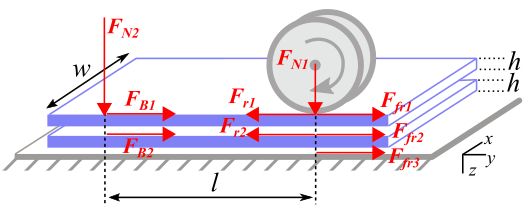}
    \caption{Schematic drawing of the physical elements involved in the problem: two layers of material laying on a table and a roller in contact with the upper one.}
    \label{fig:basicProblem}
\end{figure}

\subsection{Mathematical description of the problem} \label{sec:mathematicalModel}

If we explore the simplest case using a roller fingertip, and assuming there are only 2 layers, the problem is depicted in Fig.~\ref{fig:basicProblem}. We first want to move the top layer without moving the lower layer when the roller rotates. Following the notation in the figure, when the roller is pressed against the table with a normal force $F_{N1}$ and rolls, it generates a traction force $F_{r1}$ on the first layer that will be as big as the friction of the roller with the first layer, $F_{fr1}$. The resistant force is the friction between both layers $F_{fr2}$.
We neglect the weight of the layers so that the same normal force is transmitted to the lower layer at the same contact point. Therefore, the traction force acting on the second sheet is as much as the friction force between both layers $F_{fr2}$, and the resisting force will be the friction of the lower layer and the table $F_{fr3}$. If no other forces are applied, we can say that the top layer will be moved and the lower layer will stay still if
\begin{equation}\label{eq:successConditionWithoutFingers}
    F_{fr1} > F_{fr2} \text{ and } F_{fr2}<F_{fr3}. 
\end{equation}
As the normal force is the same in all layers, inequalities~\ref{eq:successConditionWithoutFingers} depend only on the friction coefficients of the materials. One can design a roller with a high friction coefficient that will make the first condition hold; however, tables usually have rather small friction coefficients; therefore, the second condition is more difficult to satisfy.

To solve this issue, we can apply a sufficiently large second normal force $F_{N2}$ at a distance $l$ from the roller contact. This second normal acts as a holding point that prevents sliding of both layers, and the system becomes equivalent to that of a buckling slender column. That is, with the rotation of the roller, the roller force needs to overcome not only the friction but also the Euler's buckling critical load. According to Euler–Bernoulli theory, a pin-ended column under the effect of buckling load will bend at the critical force of
\begin{equation}\label{eq:eulerBuclking1}
    F_B = \frac{\pi^2 E I}{ l^2}
\end{equation}
at the first mode of buckling, where $E$ is the Young's modulus of the layer material, and $I$ is the second moment of area of the cross section of the layer. The term $E I$ is known as flexural rigidity. $E$ represents the stiffness of the cross-section due to its material and $I$  represents the stiffness of a beam cross-section due to its geometry. In our case, we can compute  $I$ following the formula of the second moment of inertia for bending about the $x$ axis for a rectangular cross-section, that is
\begin{equation}
    I_x = \frac{w h^3}{12}
\end{equation}
where $w$ is the width of the layer and $h$ its thickness, following notation in Fig. \ref{fig:basicProblem}. Substituting in equation \ref{eq:eulerBuclking1}, we have that the critical load to overcome is
\begin{equation}\label{eq:eulerBuclking2}
    F_{Bi} = \frac{\pi^2 E_i w h^3}{12 l^2},
\end{equation}
where $i$ refers to the layer number, so that $E_i$ is the Young's module of the material of each layer.

In this new case, the first layer will buckle if the overall traction $F_{fr1}-F_{fr2}$ exceeds the critical load $F_{B1}$ and the second layer will stay still if its overall traction force $F_{fr2}-F_{fr3}$ cannot overcome the buckling critical load $F_{B2}$. Therefore, when the normal force at distance $l$ is applied, the new conditions to allow layer separation are as follows:
\begin{equation} \label{eq:successConditionWithFingers}
   F_{fr1}-F_{fr2} > F_{B1}  \text{ and } F_{fr2}-F_{fr3}< F_{B2}. 
\end{equation}
If we compare this with the previous condition without a holding force in equation~\ref{eq:successConditionWithoutFingers}, we can see that now, even if the friction with the table is low, we can use the buckling critical load to ensure that the second layer does not move, assuming the holding force $F_{N2}$ is large enough to prevent motion at contact.

\begin{figure}[tb]
    \includegraphics[height=0.3\linewidth]{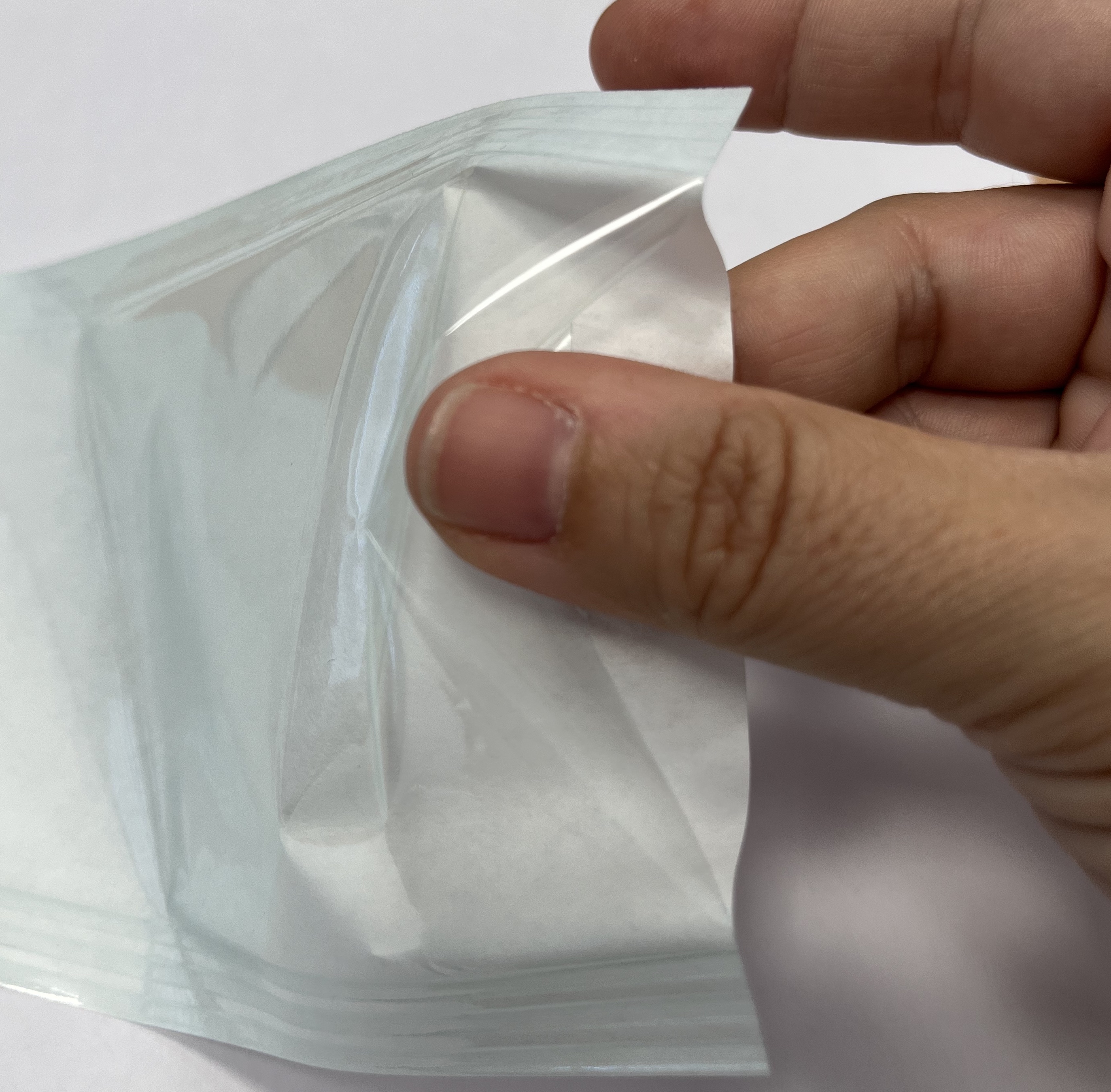}
    \includegraphics[height=0.3\linewidth]{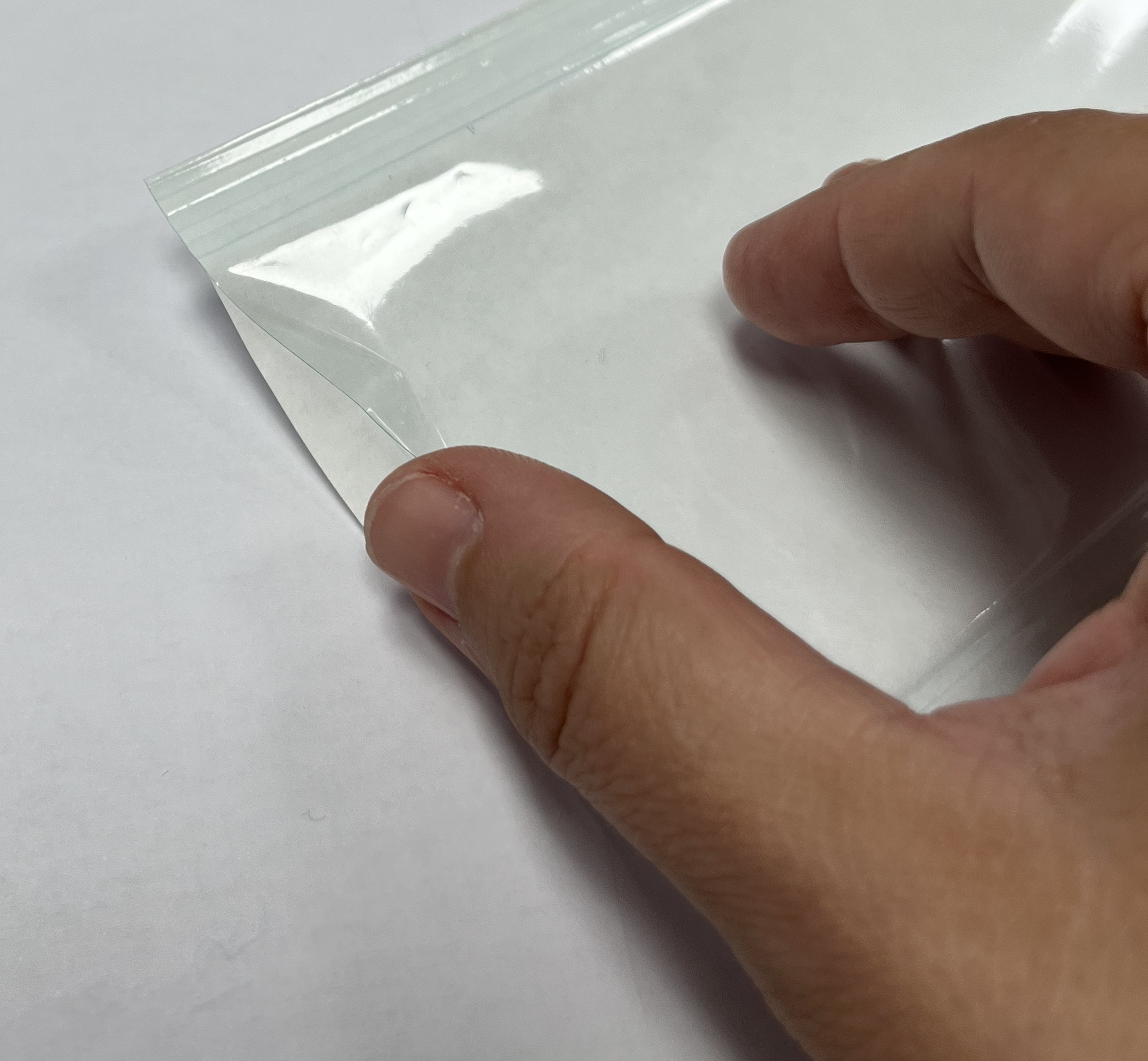}
    \includegraphics[height=0.3\linewidth]{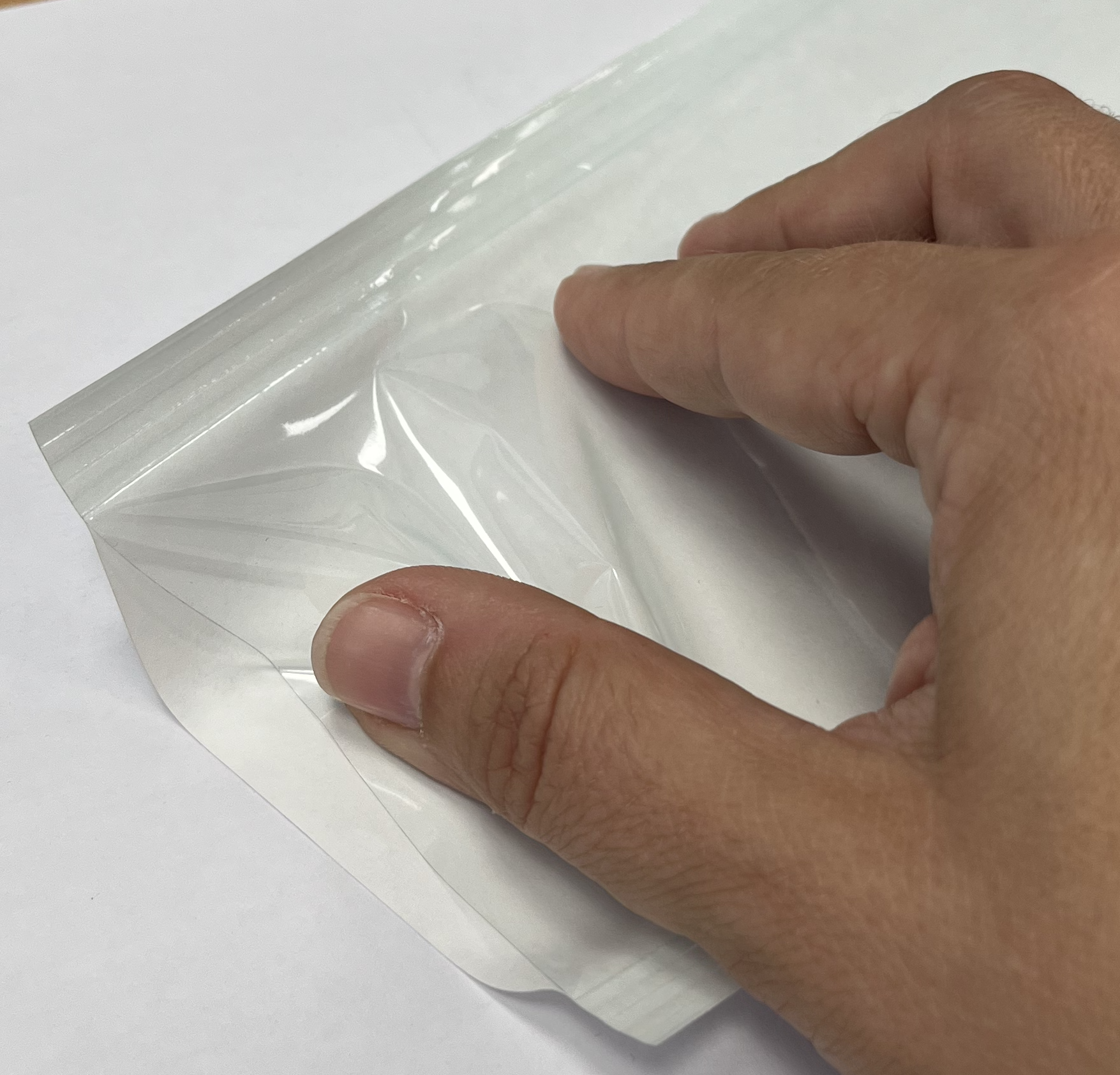}
    \begin{tabularx}{\linewidth}{@{}*3{>{\centering\arraybackslash}X}@{}}
        (a) & (b) & (c)\\
    \end{tabularx}
    \caption{Examples of a human manipulating a two layer flat pouch for layer separation: a) Rubbing against each other: grasping on both sides of the layers, moving the finger contact points one against the other to drag away one of the layers and grasp the other. b) Edge discovery: while on a table or in the air, positioning one of the fingers over the edge so that only one of the layers moves. c) Pinch-like manipulation: on a table, one of the fingers holds the layers against the table, to prevent them from moving, and the other finger performs the right amount of pressure to only drag the top layer.
    }
    \label{fig:humanExamples}
\end{figure}

It is important to note that we can increase the buckling critical load by reducing the proximity of the holding force to the roller, according to equation \ref{eq:eulerBuclking2}.

This model is a simplification of the complex interactons that take place. Previous works like~\cite{unde2024design} acknowledge, for instance, how fabrics may not satisfy the Amonton’s law that models the relationship between friction coefficient and normal forces $F_{fr}=\mu F_N$. In addition, when dealing with plastic sheets, electrostatic forces between layers may prevent any relative motion between them. However, the mathematical model, together with the observation of the task performed by humans, can inspire designs that may then be tested in real settings for evaluation, as we will do in the following sections.

\section{Gripper design} \label{sec:design}

When humans perform layer separation, several strategies are followed before one of the flaps can be grasped; see examples in Fig.~\ref{fig:humanExamples}.
Taking into account both the strategy in Fig.~\ref{fig:humanExamples}c and the theoretical analysis of the previous section, it can be seen how intuitively humans perform the normal force at a distance, allowing the buckling force to play a role to facilitate separation. Without it, most of the time we would just drag both layers at the same time. We get inspiration from the combinations of approaches Fig.~\ref{fig:humanExamples}(b and c), playing the role of the thumb with an active roller.

\begin{figure}[bt]
    \centering
    \includegraphics[width=0.9\linewidth]{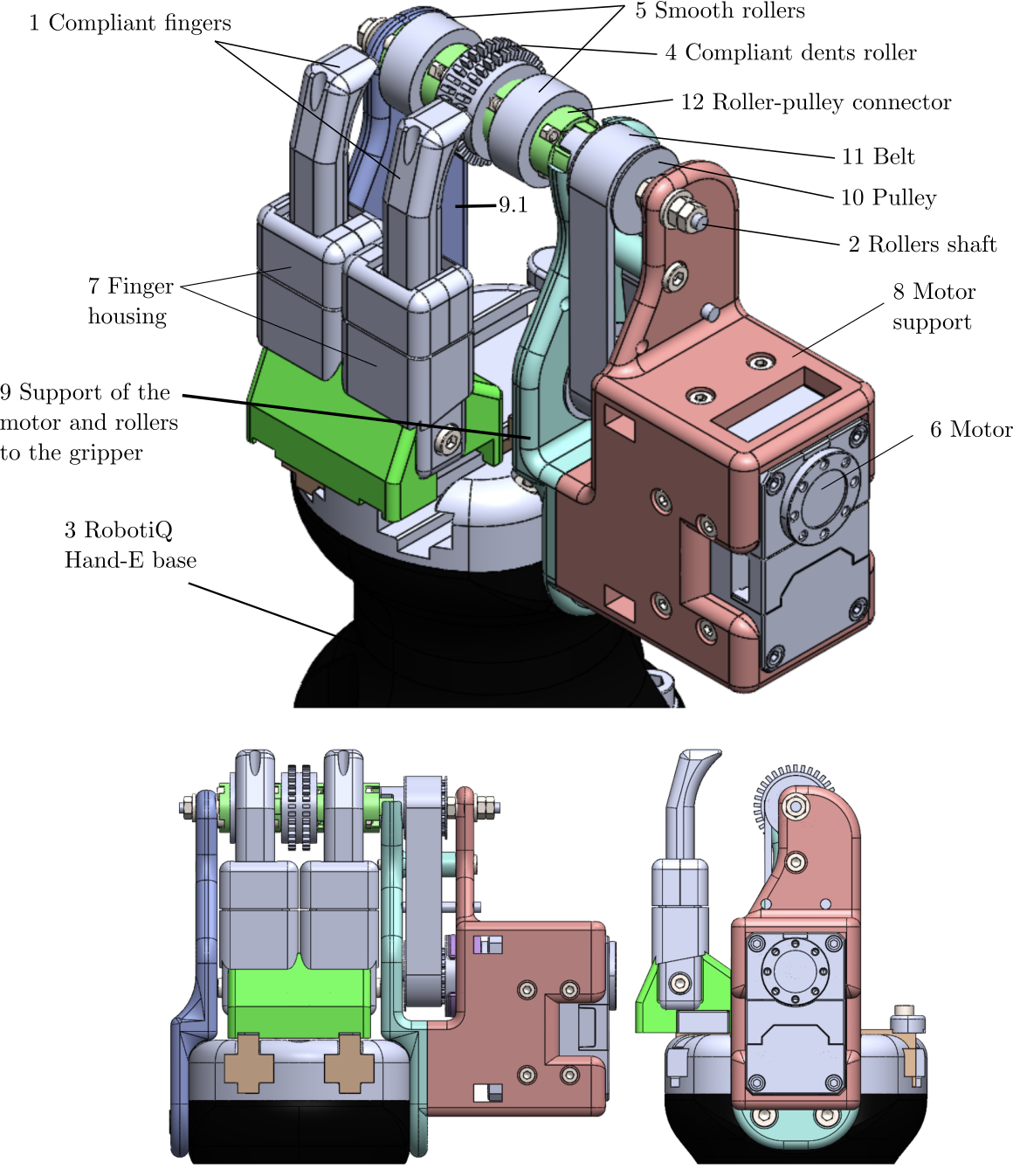}
    \caption{Our gripper can be mounted on the RobotiQ Hand-e gripper. The actuated roller structure is fixed, and the couple of holding flexible fingers can be closed against the roller using the Hand-e actuator. For reproducibility purposes, here there are some details of the main parts of the gripper: part 6 is a Dynamixel XM430-W350, part 11 is a Contitech 10/T5/150 SS polyurethane synchronous timing belt, parts 4 and 5 are made of Dragon Skin 30 silicone, as well as the tip of the fingers (part 1).}
    \label{fig:gripperDesign}
\end{figure}

\begin{figure*}[bt]
\centering
\includegraphics[width=0.9\textwidth]{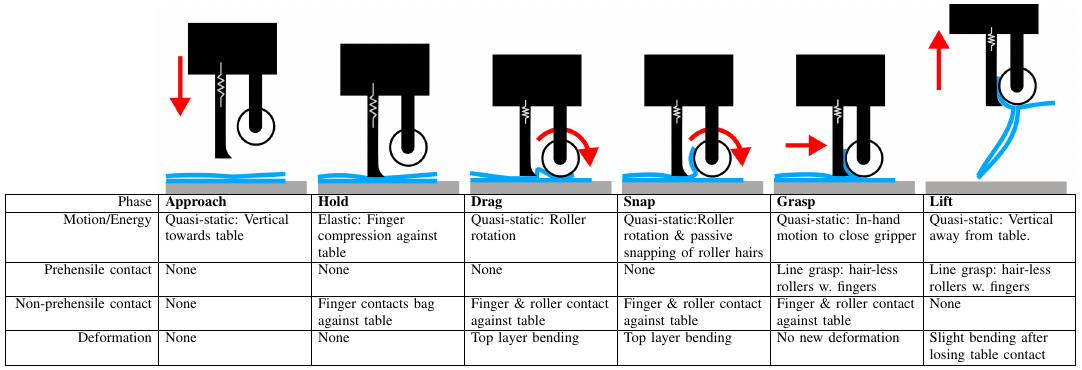}
\caption{Phases of the layer separation. In each phase we describe the interactions following the taxonomy introduced in~\cite{blanco2025TDOM}.}
\label{fig:phases}
\end{figure*}

Our gripper design can be seen in Fig.~\ref{fig:gripperDesign}. Each part is labeled and numbered to allow easy referencing, and from now on they will be referred as part $i$, for $i=1,\dots , 12$. The way the gripper is operated is depicted in Fig.~\ref{fig:phases}. 
Our design motivation is to utilize an off-the-shelf parallel gripper, such as the RobotiQ Hand-E, and adapt it to our needs. As most robotic grippers do not have the dexterity to allow a relative motion between the fingertips to rub against each other (strategy Fig.~\ref{fig:humanExamples}a), not even the ones used for the majority of humanoids, a common solution is to use active rollers as fingertips. 
Our solution consists in first, canceling one of the proprietary gripper's fingers; second, adding a new fixed finger with a set of active rollers as a fingertip (parts 4 and 5) and third, substituting the remaining proprietary parallel finger by two compliant fingers (part 1) that solve two tasks: holding against the table and grasping (2nd and 5th phase in Fig. \ref{fig:phases}). 

There are several design choices that are not trivial: 

\textbf{Compliant double fingers:} First, instead of using one finger, we use two (part 1). This is because when the flap of the bag is grasped to be opened, having a single grasping point led to easy undesired rotations that resulted in breaking the bag when opening. Instead, having a linear grip along the flap proved to be more stable to maintain the orientation of the bag while opening it. Finally, in order to securely contact the table to hold the bag, we added springs on the base of the fingers, inside each finger housing (part 7).

\textbf{Triple roller with single shaft:} Our design divides our roller into 3 rollers: two smooth rollers (part 5) and one dented-roller (part 4), all activated with a single shaft. The two smooth rollers are there to firmly grasp the flap against the double fingers (part 1).

\textbf{Dented-roller:} The roller that performs the drag of the top layer is the central one (part 4). It has two main characteristics: first, it needs to be located at the fingertip and second, it is dented. The first characteristic is needed to achieve the relative motion at the contact. To achieve this, as is common in many designs, the motor (part 6) is mounted at the base of the Hand-e gripper, and a belt (part 11) with a pulley (part 10) is used as a transfer mechanism. The second characteristic was designed through experimentation. Using smooth rollers with very high friction achieved successful dragging of the flap but required a very precise vertical positioning to avoid dragging the second layer. Adding dents or hairs introduced a passive compliant element that facilitated the first contact. In addition, we want the flap to snap to the other side of the roller to perform a secure clamping grasp instead of a pinching grasp, and the dents introduce a passive dragging element that facilitates the snapping of the layer once the edge is reached, similar to what humans do in strategy Fig.~\ref{fig:humanExamples}b. The effectiveness of this design is evaluated in Section~\ref{sec:dentRollerEvaluation}.

\textbf{Finger profile:} The geometry of the fingers is meant to adapt to the smooth rollers when the fingers are closed while in contact with the table (grasping phase in Fig.~\ref{fig:phases}). The force of the fingers against the roller must be strong so that the flap is firmly grasped and can be pulled to open the bag. In Section~\ref{sec:forceHoldingEvaluation}, we evaluate how much force it can withstand.
The gripper closes while in contact with the table, therefore, keeping the fingers flexed. Closure force maintains the fingers flexed even after removing the contact with the table (lifting phase in Fig.~\ref{fig:phases}). Once the gripper opens, the fingers recover their resting position thanks to the springs located inside the housings (part 7).

\textbf{Fixed roller, moving fingers:} The mechanism of the actuated roller is firmly attached to the gripper base (part 9) and therefore does not move against the other finger. In contrast, the double flexible fingers are mounted on the mechanical piece of the finger (green part in Fig.~\ref{fig:gripperDesign}) that moves when the proprietary gripper motor is activated, providing a linear motion that opens and closes against the roller.

In Fig.~\ref{fig:phases}, we systematically describe all phases of gripper operation using
a taxonomy for the manipulation of deformable objects introduced in~\cite{blanco2025TDOM}. The different phases are
\begin{itemize}
    \item \textbf{Approach}: Before contact, the gripper descends vertically. 

    \item \textbf{Hold}: The gripper moves until the roller touches the first layer. Before that, the first contact happens with the flexible fingers that hold the object in place.
        
    \item \textbf{Drag}: When the roller makes contact, it starts to rotate to drag the top layer.

    \item \textbf{Snap}: When the edge of the top layer is reached, the side of the flap that is in contact with the roller flips.

    \item \textbf{Grasp}: Fingers in contact with the table close against the roller to grasp the flap.

    \item \textbf{Lift}: The contact with the table is removed to move closer to the other gripper, which will then grasp the other flap and then pull to open the bag.

\end{itemize}

Our analysis includes a description of the type of motion, the prehensile and non-prehensile contacts, and the deformation that they induce. In our experiments, we see how the fingers perform a non-prehensile grasp of the bag against the table while the roller moves, and how the same fingers later perform a prehensile grasp. The motion of the roller needs to buckle the layer, that is labeled as bending deformation following the taxonomy. 

\begin{figure}[bt]
    \centering
    \includegraphics[width=0.8\linewidth]{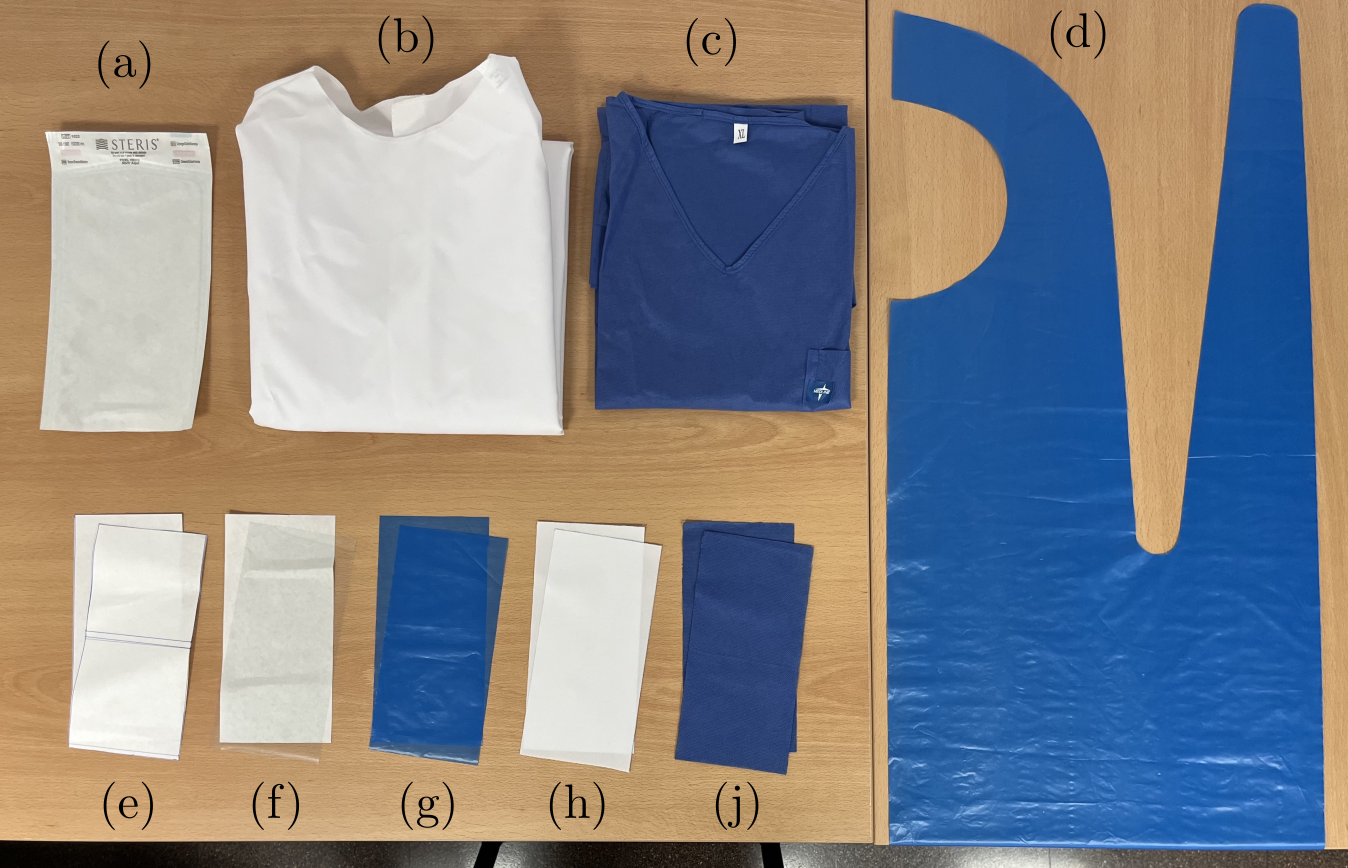}
\vspace{0.1cm}
    
    \small
    \setlength\tabcolsep{3pt}
    \begin{tabular}{c|c|c|c|c|c|c|}
                       & (e)     & (f): paper & (f): plastic  &  (g)     & (h)     & (j)    \\ \hline
       Stiffness       &  95,7\%  & 95,7\%     &  54,4\%      &  27,8\%  & 38\%  & 29,1\% \\ \hline
       Friction c.     &  0,26   &  \multicolumn{2}{c|}{0,36}    &  0,38    &  0,61  & 0,34   \\  \hline
    \end{tabular}
    \caption{Set of material tested. (a) is a 15x30 cm empty bag used in the hospital to seal sterilized material, (b) and (c) two different medical coats, (d) a plastic medical apron folded in half. (e-j) 10x21 cm two layers of cut out materials from the objects above: (e) two layers of the paper layer from the bag object in (a), (f) a plastic-paper layers from (a), (g) cut from (d), (h) cut from (b), and (j) cut from (c). In the table we show indicative measures of stiffness measured following ~\cite{garcia2024standardization} and friction coefficients measured following~\cite{oriol2023gripper}.}
    \label{fig:objects}
\end{figure}

\section{Evaluation}\label{sec:evaluation}
We performed different evaluations to test the gripper performance and study the differences with different materials. In Fig.~\ref{fig:objects}, we show a summary of all the materials tested, although most of the tests to open sealed bags were carried out with the official bags used in the hospital to seal the sterilized material (Fig.~\ref{fig:objects}a).

\subsection{Buckling force and finger design} \label{sec:firstEvaluation}
\begin{figure}[bt]
    \centering
    \includegraphics[width=\linewidth]{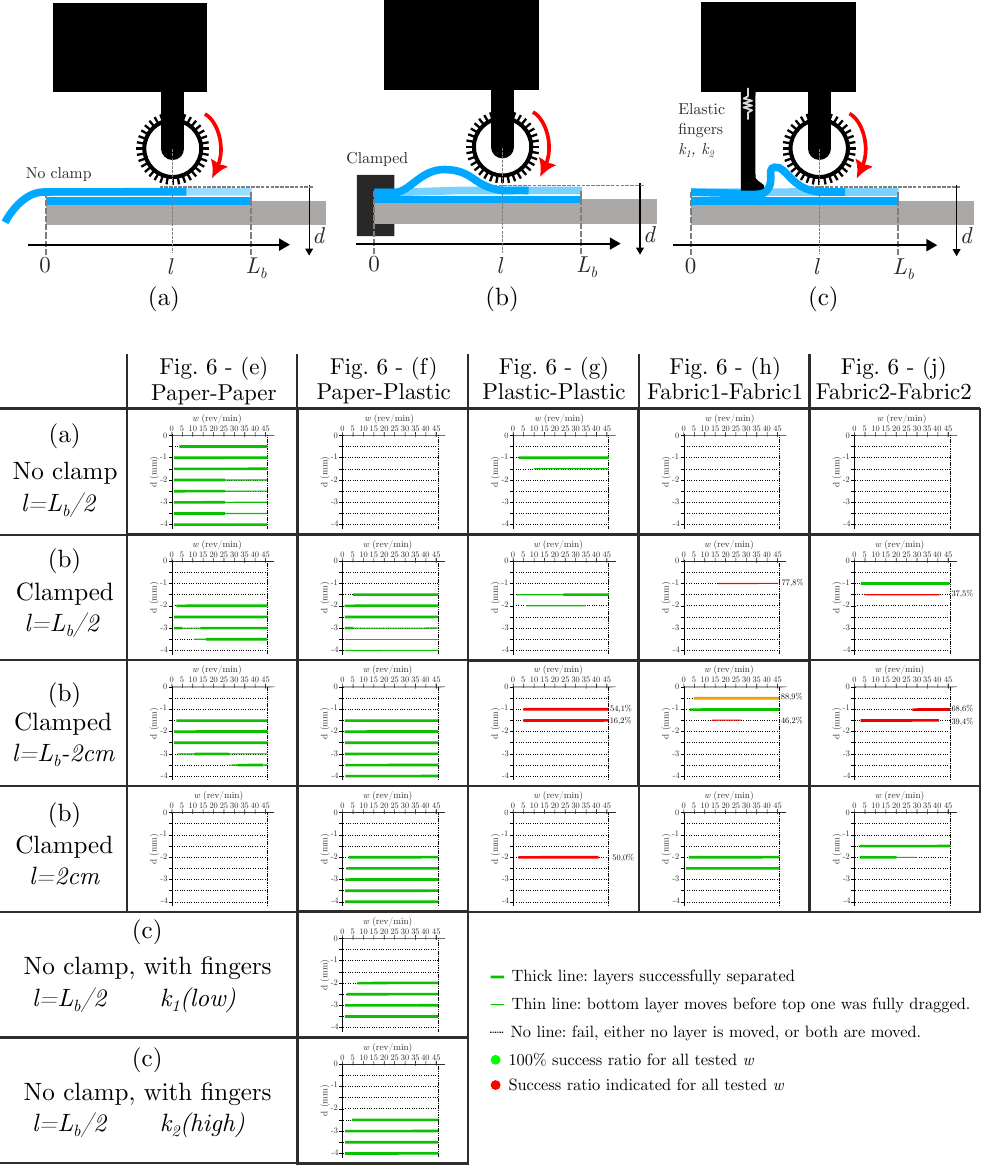}
    \caption{Results of the theoretical buckling force evaluation of the dented-roller in different clamping scenarios: a) two unclamped layers of material, b) both layers strongly clamped to the table and c) both layers hold by two compliant fingers. Each cell in the table shows the experiments with the different penetration distances (so, different normal forces $F_N$) and increasing roller angular velocities (i.e., increasing roller force $F_r$). Each column correspond to different materials using the cut-outs from Fig.~\ref{fig:objects}(e-j).}
    \label{fig:experimentalResults1}
\end{figure}

We performed a first set of evaluations to understand how the theoretical model applies to the practical problem. The summary of results can be seen in Fig.~\ref{fig:experimentalResults1}. The first four rows test only the roller (without the fingers) in two scenarios: with the two layers laying on the table (first row) and with both layers clamped to a table so that they could not move, meaning the buckling force starts playing a role.

The robot was operated in position control and we could operate the roller at different velocities, thus we repeated each experiment at different penetration positions, from 0 (where there was visual contact but the layer was still not dragged) to -4mm, corresponding to increasing normal forces $F_N$. For each normal force, we repeated the experiment 40 times at increasing angular roller velocities, from 1 to 45 rev/min (the minimum and maximum speeds allowed by the motor).

As analyzed in Section~\ref{sec:mathematicalModel}, without clamping (first row), the success fully depends on the friction coefficients. Since we made the roller of silicon, the friction of the roller with the first layer (labeled $F_{fr1}$ in Fig.~\ref{fig:basicProblem}) is easily the larger one always. However, we cannot control the friction between the layers and the table to hold Eq.~\ref{eq:successConditionWithoutFingers}. Success occurs only with the paper-paper and the thin apron plastics (Plastic-Plastic). Note that, for the Plastic-Paper (Fig.~\ref{fig:objects}f), the paper in contact with the table is the same as in the first column, but the friction between layers is larger with the plastic. With the thin plastic layers from the apron (Fig.~\ref{fig:objects}g), only one normal force was successful, meaning that one needs to perform just the right amount of normal force. The second row adds the clamp, and then for the paper-paper case we still get successful separation but with higher normal forces, and for the plastic-paper we can now successfully perform the separation of layers, thanks to the buckling force playing a role. For the apron plastic-plastic material (3rd column), we still can perform the task but for an even fewer rank of parameters (a bit higher normal force and higher roller velocity). For fabrics, the white fabric (Fig. \ref{fig:objects}h) which is more rigid and with more friction is still quite unstable, while the blue medical coat fabric (Fig.~\ref{fig:objects}j) becomes successful with a light normal force.

The next two rows display results for different locations of the roller contact, that is, different $l$ parameters for the buckling force (Eq.~\ref{eq:eulerBuclking2}). A larger $l$ (third row) means that the buckling force is lower, but it is also less the amount of material that needs to be dragged (material on the right-hand side of the roller). Thus, we see a bit less normal force is enough, and for the plastic-paper (column 2) a wider range of normal forces succeed at the task. When $l$ was low (4th row), the buckling force is larger, but also the amount of material to be dragged is very large. We could not succeed with the paper-paper because the needed buckling force became too large. We required higher normal forces for the other cases with respect to the case before. Instead, for the fabrics that was the most successful case, although still at a very narrow span of normal forces.

\begin{figure}[tb]
    \centering
    \includegraphics[width=0.9\linewidth]{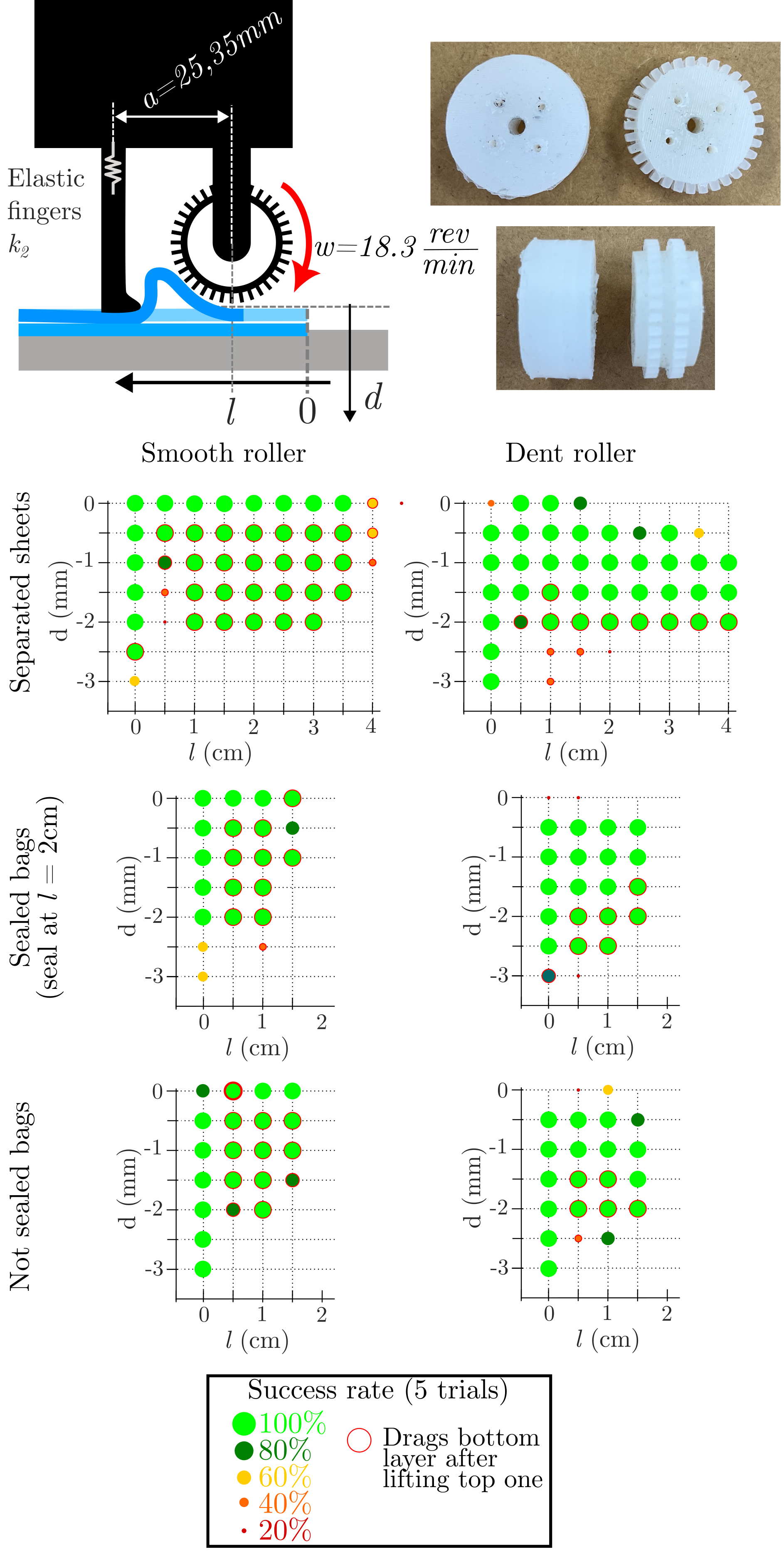}
    \caption{Results of comparing success rate of the snapping step from Fig. 4 with two different rollers: one with smooth surface and the other dented. The red circle over indicates that roller lifts the top layer successfully, but then it start lifting the bottom layer as soon as the top one has been lifted.}
    \label{fig:exp_dents_nodents}
\end{figure}

The conclusion of the first four rows is that clamping the layers allows the task to be successful for all materials. The compliant fingers (Fig.~\ref{fig:gripperDesign}~part 1) are designed to hold the object on the table acting as a clamp as long as they perform enough normal force to hold the object in place. This is demonstrated in the experiments in the last two rows (Scenario $c$ in Fig.~\ref{fig:experimentalResults1}).
Note that clamping with the fingers allows the term $l$ from the buckling force that appears in Eq.~\ref{eq:eulerBuclking2} to be substituted by the pre-opened fingers distance, $a$ in Fig.~\ref{fig:exp_dents_nodents}. This means that we can increase the buckling force by pre-closing the gripper independently of the roller contact location.

The reason why no major difference is seen when buckling forces appear for fabrics is that such forces are generally low because fabrics are less stiff, and therefore this does not influence so much. 

\subsection{Dented-roller} \label{sec:dentRollerEvaluation}
The gripper design allows to quickly change the roller into different surfaces and textures, allowing to compare performance depending on the roller surface for the exact same task. 

We compared the influence of having dents on the roller surface, with the results shown in Fig.~\ref{fig:exp_dents_nodents}. Compared to the previous experiment, we focused on the paper-plastic case (Fig.~\ref{fig:objects}f), as these are the materials used for most of the medical packages for the operating room preparation. We also fixed the experiments at an angular velocity $w=18.3 rev/min$, as this gave the best performance in the previous experiments, and the fingers pre-opened at a distance $a=25,35mm$ with high finger elasticity ($k_2$). 
In the graphics, we show the success rate over 5 repetitions for each combination of distance from the roller contact point to the edge of the material ($l$) and the contact force (measured as the penetration distance $d$). The first row is the same task as in the last row of the previous experiments, that is, two unclamped sheets of material (Fig.~\ref{fig:objects}j). The second row corresponds to the grasping of the flaps of a sealed bag (object in Fig.~\ref{fig:objects}e). The location of the seal limits the maximum distance the roller can be placed. The last row corresponds to the same bags but without being sealed (that is, the same object as before but grasped at the opposite side where the seal is). The case is different from that of the first row because the sheets are still sealed on the lateral sides, so they are not independent sheets. That also means that if the roller contact occurs far from the edge, it will not work, as the lateral seals will prevent the separation of the layers. 

We can see that the main difference between performance with or without a dented-roller surface is the ability to perform the task without dragging the bottom layer. When this happens, we label the task as successful because the top layer has been correctly separated and lifted, and if the roller rotation is stopped at that precise moment, the task is successful. However, if the roller continues to rotate a bit more, the system also drags the bottom layer into the grasp. Stopping the roller at the correct time is challenging because the lifting time may depend not only on the location of the roller contact ($l$) and the angular velocity of the roller ($w$) but also on the properties of the materials handled. Therefore, the dented-roller is much more robust in performing the task successfully without having to correctly estimate when to stop the roller rotation.

In addition, we can also see that locating the roller contact point precisely at the edge of the layers ($l=0$) allows to correctly separate the layers with both dented and smooth rollers. That situation is similar to the human example in Fig.~\ref{fig:humanExamples}b and it works because the snapping happens from the beginning of the motion. However, note that the roller contact location has to be very precise for this case, which also means that it will be less robust when we use a perception method to locate the bag before grasping. Therefore, the case $l=0$ is much less robust than any other $l>0$.

\begin{figure}
    \centering
    \includegraphics[width=0.9\linewidth]{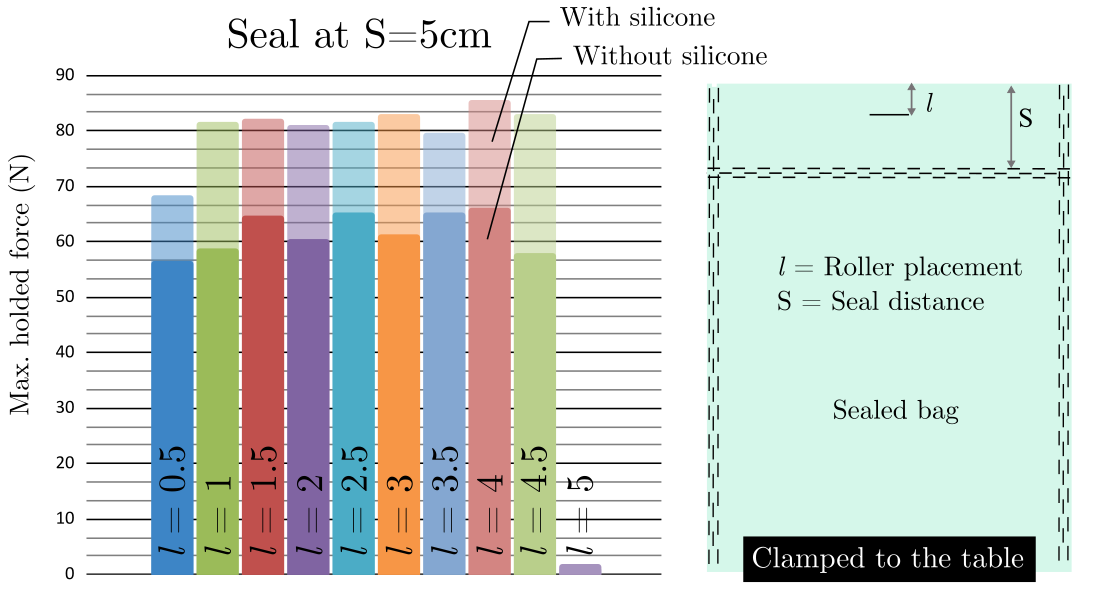}
    \caption{After the flap was separated and grasped, the robot arm moves up the bag. As the bag is clamped to the table, the motion up is maintained until the grip is lost, capturing the pulling forces until it detaches. The graphic shows the maximum holding force read by the UR5E wrist force sensor. We show values with the fingers covered with silicone, vs. not covered with silicone.}
    \label{fig:expeirments_force}
\end{figure}

\subsection{Grasp holding force evaluation} \label{sec:forceHoldingEvaluation}

The final goal for our gripper is to open sealed bags, therefore the flap grasp force must be strong enough to peal open the seals. 
In this section, we show the results of the amount of force the fingers could hold after the flap is grasped, summarized in Fig.~\ref{fig:expeirments_force}. To do so, we designed an experiment where the sealed bags are clamped on the table at the location shown at the figure. We perform the flap grasp procedure shown in Fig. \ref{fig:phases}. During the last lifting phase, as the bag is clamped to the table, the robot arm keeps pulling up until the grasp is lost, and we record the maximum pulling force before the grasp is lost.
The grasp is performed at different roller contact locations with respect to the edge, and we can see how even very close to the edge, that is, with a very short flap, the grasp is strong enough to withstand more than 55N.

Finally, we also show results with the fingers coated in silicone versus not-silicone-coated fingers. As expected, the silicon-coated fingers can withstand 15-20N more pulling force, although the force without silicone is already quite large because the side rollers (part 5 in Fig.~\ref{fig:gripperDesign}) are silicone (with a smooth surface). Note that grasping against the dent roller would result in less force withstanded and a quick degradation of the dents with the long-term use of the gripper.

\subsection{Opening sealed bags} \label{sec:openSealedBags}

The scope of this paper focuses primarily on a gripper design that enables the pre-grasping phase required for opening sterile pouches. To demonstrate the gripper's full capabilities beyond flap grasping, we implemented an open-loop procedure using a dual-gripper configuration to execute an opening trajectory inspired by clinical aseptic protocols for preventing internal contamination. While a robust methodology for pouch opening is beyond the scope of this work, it is essential to demonstrate that the achieved grasp enables successful task execution. As previously stated, the dual-finger and dual-roller architecture facilitates a linear grasp that securely constrains the flap.

Following the initial grasp of the first flap by the right gripper, the pouch is elevated, and the left gripper's fingers are positioned between the grasped and free flaps. Subsequently, the left gripper partially closes while the rollers are activated to move the flap further into the grasp as the gripper rotates into the pulling configuration. Upon reaching the target position, the left gripper completes the grasp, and both end-effectors translate in parallel, opposing directions to open the seal. This open-loop procedure was evaluated over 30 trials using fully sealed pouches and 30 trials using pouches with pre-opened side seals. Experimental results are summarized in Table \ref{tab:experimentOpeningBags}; the complete procedure is demonstrated in the supplementary video available on the project website\footnote{Proj. website: \scriptsize\url{https://sites.google.com/view/rollinggripper}}.

\begin{table}[tb]
    \centering
    \caption{Opening sealed bags results}    \label{tab:experimentOpeningBags}
        \includegraphics[width=0.9\linewidth]{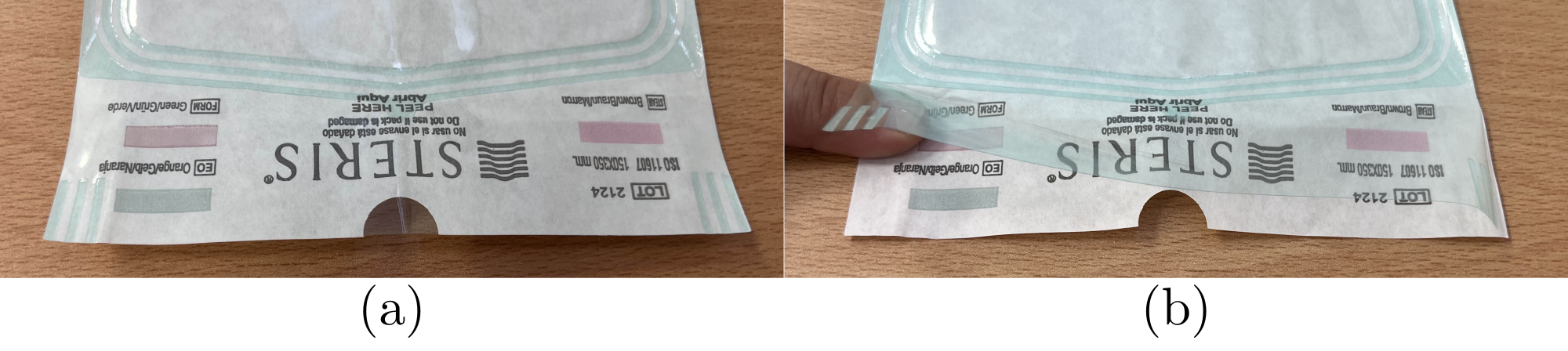}
    \begin{tabular}{c|c}
        \textbf{Object} & \textbf{Success rate} \\ \hline
        (a) Fully sealed bags & 29/31 (93,55\%)  \\ \hline
        (b) Pre-opened lateral seal & 30/31 (96,77\%) \\ \hline
    \end{tabular}
\end{table}

There are several possible causes of failure. If the flap is not well grasped (for instance, if the snapping phase does not occur), the grasp can be lost when pulling. However, the most common failure case is when the paper breaks.
For the case where the pouches are fully sealed, which means that there are lateral seals until the top of the flaps, after the flap is grasped, it bends in a way that can lead to breaking the paper when it is pulled to open. Instead, pre-opening the lateral seals (a step that nurses often do before pulling to open) prevents this bending, favoring grasp success. In any case, our results show that the success rate is very high.

\section{Discussion and Conclusions}
 
This work presents, to our knowledge, the first robotic gripper specifically designed to perform the complex task of opening sealed sterile pouches. The proposed design combines an active dented-roller fingertip with compliant fingers and has demonstrated robust performance in separating and grasping thin, flexible layers of deformable materials, such as paper, plastic or textile.

The mathematical model was useful to guide the gripper’s design, but capturing the full complexity of the interactions between the fingers, roller, and materials is very challenging. Therefore, we have exploited rapid prototyping to iteratively test several versions of the gripper capabilities in real settings. Our experiments show that the gripper is robust for the tested hospital sealed pouches but also to what extent its performance depends on the applied normal force. For materials such as plastic aprons, which were successfully separated only for a very narrow range of normal forces, highly accurate position and/or force sensing will be essential for fully autonomous operation.

Beyond opening sealed bags, the gripper has demonstrated versatility in other potential applications. For instance, in~\cite{BlancoPreDressing2025}
it was used to grasp the necks of folded medical gowns, and it can readily perform pinch grasps of single layers. By changing the central roller coating, a wider range of materials and tasks could be addressed.

Future iterations will transition from current open-loop constraints to a unified closed-loop control strategy that encompasses the entire pipeline, from pouch opening to sterile content delivery. Key objectives include implementing sensor-based path planning to improve autonomy and performing rigorous comparative benchmarking against human cycle times and robotic baselines to improve the system's speed and to validate system utility in clinical environments.

\looseness=-1
\bibliographystyle{ieeetr}
\bibliography{clothManipulation}

\end{document}